\documentclass[11pt]{article}

\usepackage[final]{acl}
\usepackage{makecell}
\usepackage{tabularx}
\usepackage{array}
\newcolumntype{L}{>{\raggedright\arraybackslash}X}

\usepackage{algorithm}
\usepackage{algpseudocode}
\usepackage{tablefootnote}

\newcommand\blfootnote[1]{%
  \begingroup
  \renewcommand\thefootnote{}\footnote{#1}%
  \addtocounter{footnote}{-1}%
  \endgroup
}
\usepackage{listings}
\usepackage{subcaption}
\usepackage[most]{tcolorbox}
\usepackage{times}
\usepackage{amsmath}
\usepackage{latexsym}
\usepackage{booktabs}
\usepackage{multirow}   
\usepackage[T1]{fontenc}

\usepackage[utf8]{inputenc}

\usepackage{microtype}

\usepackage{inconsolata}

\usepackage{graphicx}

\title{SuperValid: Capability-Aligned OOD Validation \\ for Generalizable Downstream Scaling}

\author{
  \textbf{Quanen Sun$^{*\dagger}$}, 
  \textbf{Changxin Tian$^*$}, 
  \textbf{Ke Shi$^*$}, 
  \textbf{Cai Chen}, 
  \textbf{Cunyin Peng}, 
  \\
  \textbf{Jia Liu}, 
  \textbf{Kunlong Chen}, 
  \textbf{Zhiqiang Zhang$^\ddagger$}, 
  \textbf{Jun Zhou$^\ddagger$}
\\
  Ant Group
\\
  \texttt{co2penguin@gmail.com}, 
  \texttt{\{tianchangxin.tcx,keshi.sk,lingyao.zzq\}@antgroup.com} 
}

\begin{document}
\maketitle
\blfootnote{$^*$Equal contribution.}
\blfootnote{$^\dagger$Work done during an internship at Ant Group.}
\blfootnote{$^\ddagger$Corresponding author.}
\begin{abstract}

Scaling laws guide large language model training by relating compute to cross-entropy loss, and recent work further extends them to predict downstream benchmark performance. However, prior approaches face generalization limitations from two aspects: focusing on benchmark-level performance introduces scenario-specific artifacts, while relying on IID validation loss fails to track capability improvements when training distributions vary. In this work, we argue that downstream scaling should be studied at the \emph{capability level}, which captures shared skill factors across related tasks while abstracting away benchmark-specific noise. We propose \textbf{SuperValid}, a framework that synthesizes OOD (out-of-distribution), capability-aligned validation data by distilling core concepts from benchmarks within a capability domain and expanding them into diverse, knowledge-rich texts. Extensive experiments spanning 16 benchmarks grouped into 6 capability domains show that SuperValid loss exhibits strong and stable correlation with downstream performance across models of different architectures, scales, and training data distributions. As a training-free metric computable during training without benchmark evaluation, SuperValid enables effective model selection, early stopping, and scaling decisions.

\end{abstract}

\section{Introduction}

\begin{figure}[htbp]
\centering
  \begin{subfigure}[b]{0.32\linewidth}
        \centering
        \includegraphics[width=\textwidth]{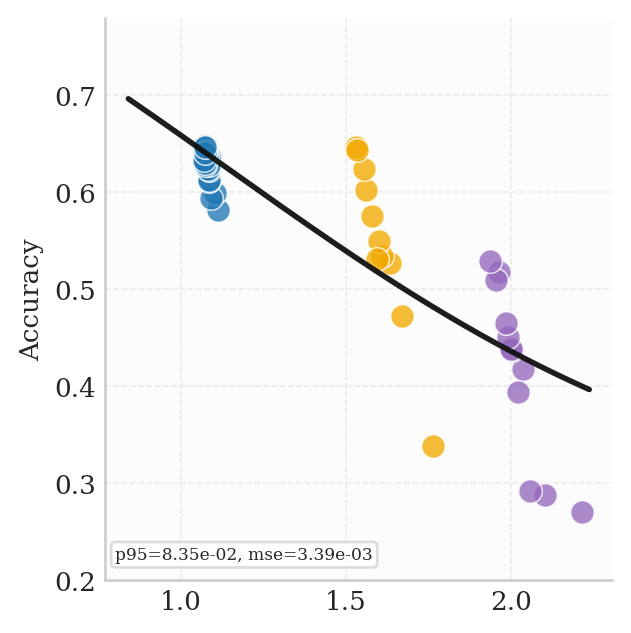} \hfill
        \vspace{-13pt}
        \captionsetup{font=scriptsize}
        \caption{IID Loss} 
        \label{fig:iid_loss}
  \end{subfigure}
  \begin{subfigure}[b]{0.32\linewidth}
        \centering
        \includegraphics[width=\textwidth]{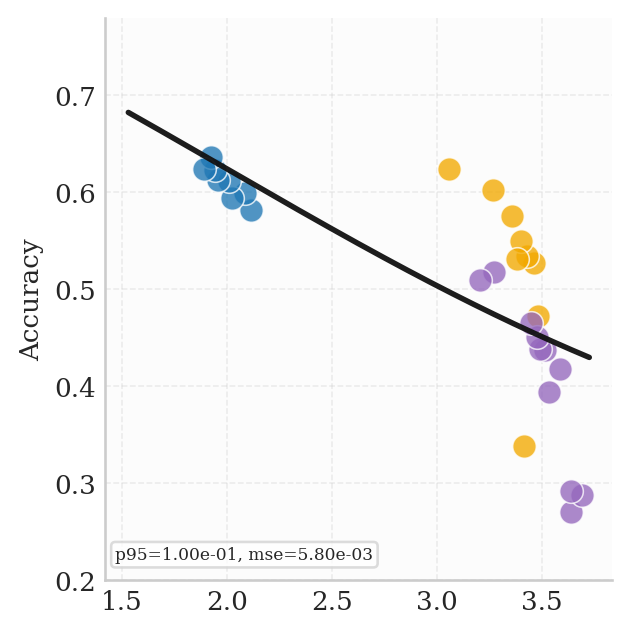} \hfill
        \vspace{-13pt}
        \captionsetup{font=scriptsize}
        \caption{Weighted IID Loss} 
        \label{fig:weighted_loss}
  \end{subfigure}
  \begin{subfigure}[b]{0.32\linewidth}
        \centering
        \includegraphics[width=\textwidth]{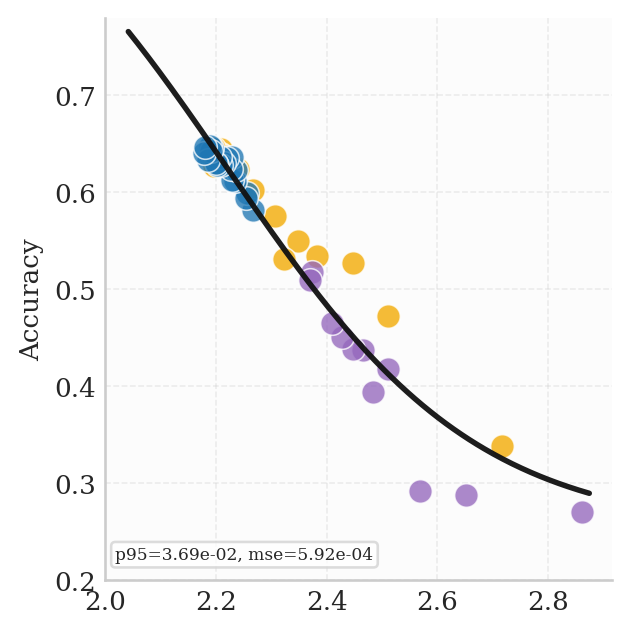} \hfill
        \vspace{-13pt}
        \captionsetup{font=scriptsize}
        \caption{SuperValid Loss} 
        \label{fig:supervalid_loss}
  \end{subfigure}
  \caption {Comparison of loss-performance correlation on CEval~\citep{huang2023ceval} across three methods: IID Loss, weighted IID Loss~\citep{ge2025capability}, and SuperValid (ours). Data points in different colors denote metrics collected from diverse training scenarios (e.g., varying architectures or datasets), while the solid black line represents the fitted sigmoid curve. SuperValid exhibits superior cross-scenario generalization.}
  \label{fig:compareIID}
\end{figure}

\begin{figure*}[t]
\centering
 \includegraphics[width=0.97\linewidth]{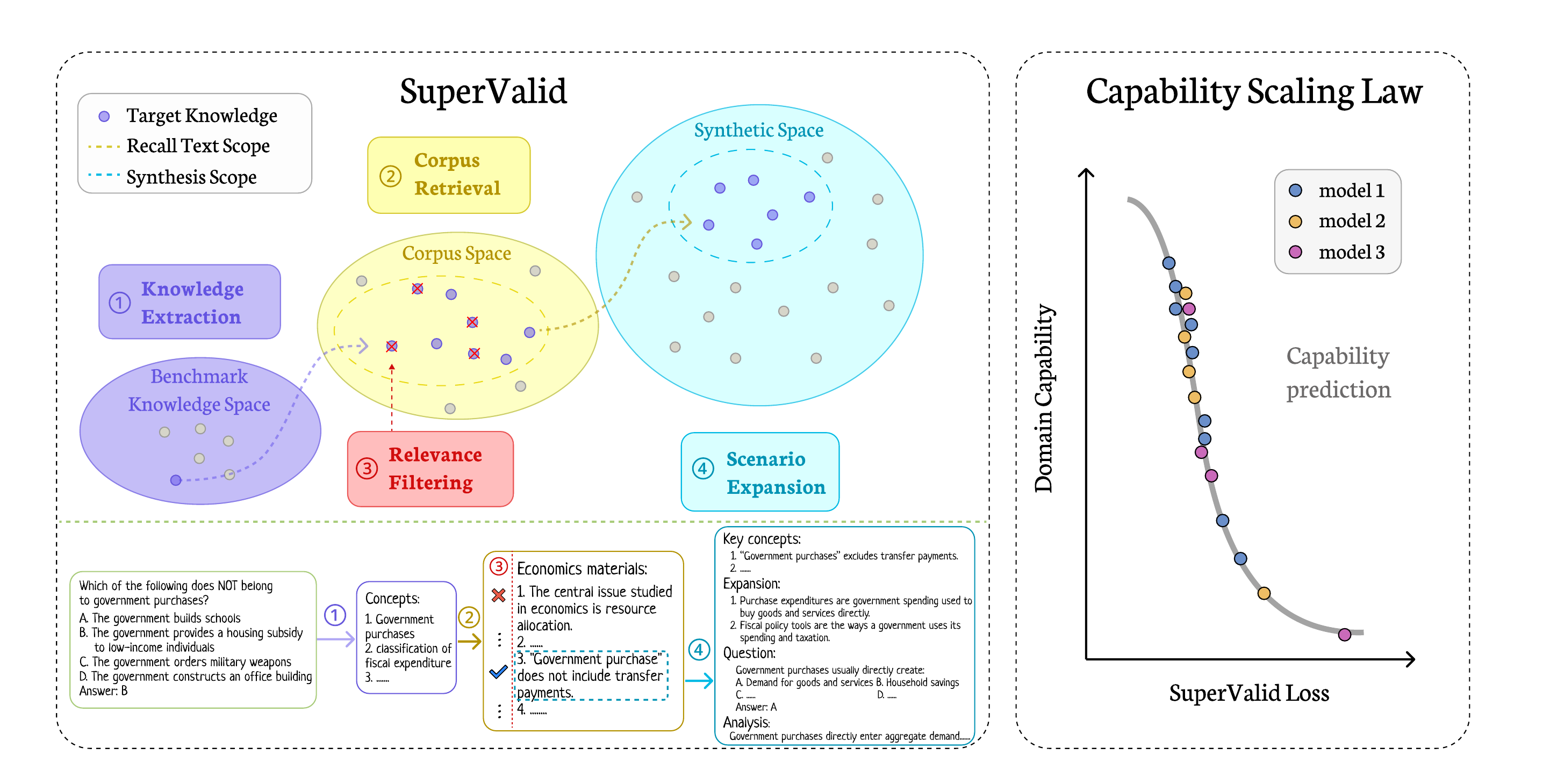} \hfill

  \caption{Overview of the SuperValid framework. Each benchmark text sample is first transformed into knowledge components via knowledge extraction, which are then used to retrieve relevant corpus text from the knowledge domain. After LLM-based filtering, the selected text is rewritten into knowledge-expanded validation samples. The loss on these samples serves as a domain capability indicator, enabling capability-level scaling law analysis. An economic example is shown for illustration.}
  \label{fig:supervalidFramework}

\end{figure*}

Large Language Models (LLMs) achieve strong performance across diverse downstream tasks, yet training them remains computationally expensive~\citep{kaddour2023challenges}. To guide efficient scaling, prior work on scaling laws~\citep{bahri2024explaining, hoffmann2022training, kaplan2020scaling, muennighoff2023scaling} has established empirical relationships between compute (FLOPs), model size, and validation loss on IID held-out data. More recently, efforts have been made to bridge validation loss and downstream benchmarks, enabling accurate prediction across a variety of tasks~\citep{gadre2025language,ruan2024observational,ge2025capability}. 

Despite their effectiveness, prior approaches face generalization limitations stemming from two key design choices: \textbf{prediction targets} and \textbf{loss computation}. 
First, existing work typically focuses on benchmark-level downstream performance~\citep{zhang2024collaborative,ge2025capability}. Since each benchmark is an entangled metric that conflates domain knowledge and format adaptation, overemphasizing benchmark-level metrics risks introducing scenario-specific artifacts such as invariant question templates and limited knowledge coverage~\citep{liang2022holistic}. Second, prior work relies on IID validation loss derived from either training or benchmark data~\citep{owen2024predictable}. However, scaling curves fitted on such IID loss fail to track underlying capability improvements, particularly when training distributions vary~\citep{liu2023same}. These two limitations jointly degrade generalization in cross-architecture and cross-data-mixture prediction settings (refer to Figure~\ref{fig:compareIID}), substantially reducing practical value in industrial training scenarios such as ablation studies over diverse data mixtures and architectures. 

To address these challenges, we argue that downstream scaling should be studied at the \emph{capability level}, an implicit variable underlying various downstream tasks that is free from benchmark-specific artifacts. Specifically, a domain-specific capability captures the shared skill factors required by a family of related tasks, explaining their common variance by extracting core concepts and skills while abstracting away scenario-specific noise. Since domain capabilities are agnostic to model-specific supervision, they provide a general indicator applicable across architectures, scales, and training data mixtures. To further decouple from the training distribution, we operationalize such capability indicators by synthesizing OOD (out-of-distribution) validation data for each domain, yielding a powerful and easy-to-use observation metric. 


Building on this insight, we propose \textbf{SuperValid}, a novel framework that constructs capability-aligned validation sets from synthetic OOD data, whose loss serves as a faithful proxy for domain-specific capabilities and a general instrument for predicting downstream task performance. Concretely, SuperValid distills capability-relevant textual content into knowledge components and expands them into capability-aware general text carrying diverse semantics.
We validate SuperValid through extensive experiments spanning 16 benchmarks grouped into 6 capability domains, evaluated on various open-source models from different families. Results show that SuperValid loss exhibits a strong and stable correlation with downstream performance across models of different architectures, scales, and training data distributions, achieving an average mean square error of 1.51e-3 across all settings. Moreover, this correlation remains highly robust under changes in data mixtures, indicating that the SuperValid synthesis pipeline can produce capability-aware validation data that generalizes to models with arbitrary architectures and data compositions. Our contributions are as follows.

\begin{itemize}
    \item \textbf{Conceptual Framework.} We reformulate downstream validation as capability measurement and propose SuperValid to synthesize OOD, capability-aligned validation data. 
    \item \textbf{Empirical Findings.} SuperValid loss strongly correlates with downstream performance across diverse models, regardless of data mixtures, scales, or architectures.
    \item \textbf{Practical Impact.} SuperValid loss can be computed during training without benchmark evaluation, enabling effective model selection, early stopping, and scaling decisions.
\end{itemize}




    
    
    


\section{Methodology}
As illustrated in Figure~\ref{fig:supervalidFramework}, our proposed \textbf{SuperValid} is a data synthesis pipeline that leverages LLMs to generate capability-aware OOD validation data without benchmark-specific artifacts. Next, we first formulate domain-level capabilities as the prediction target (\S\ref{sec:domainCap}), then describe the synthesis pipeline (\S\ref{sec:syntheticSteps}), and finally introduce the capability scaling law (\S\ref{sec:scalingLaw}).

\subsection{Preliminary: Domain Capability Formulation}
\label{sec:domainCap}

In order to define the prediction target for validation loss, we formulate model evaluation at the level of \emph{domain-specific capabilities}.
Let $\mathcal{D}=\{d_1,\ldots,d_K\}$ denote a set of capability domains (e.g., knowledge, mathematics, coding, reasoning).
For each domain $d_k$, we assume an underlying capability variable $c_k(M)$ that characterizes model $M$'s general proficiency in that domain, beyond any specific task format or benchmark.

Since $c_k(M)$ is not directly observable, we approximate it by aggregating scores from a set of benchmarks $\mathcal{B}_k=\{b_{k,1},\ldots,b_{k,n_k}\}$ associated with domain $d_k$.
Let $s(M,b)$ denote the normalized score of model $M$ on benchmark $b$. We estimate the domain capability $\hat{c}_k(M)$ by averaging benchmark scores:
\begin{equation}
    \hat{c}_k(M) = \frac{1}{|\mathcal{B}_k|}\sum_{b\in\mathcal{B}_k} s(M,b).
\end{equation}
Aggregating multiple benchmarks within a domain mitigates the bias of any single evaluation setup and yields a more robust estimate of domain-level capability.

In this work, we instantiate $\mathcal{D}$ with six domains (knowledge, basic mathematics, intermediate mathematics, code generation, code completion, and reasoning), covering representative LLM capabilities while remaining extensible to additional domains or benchmark collections.

\begin{figure*}[t]
\centering
  \includegraphics[width=0.97\linewidth]{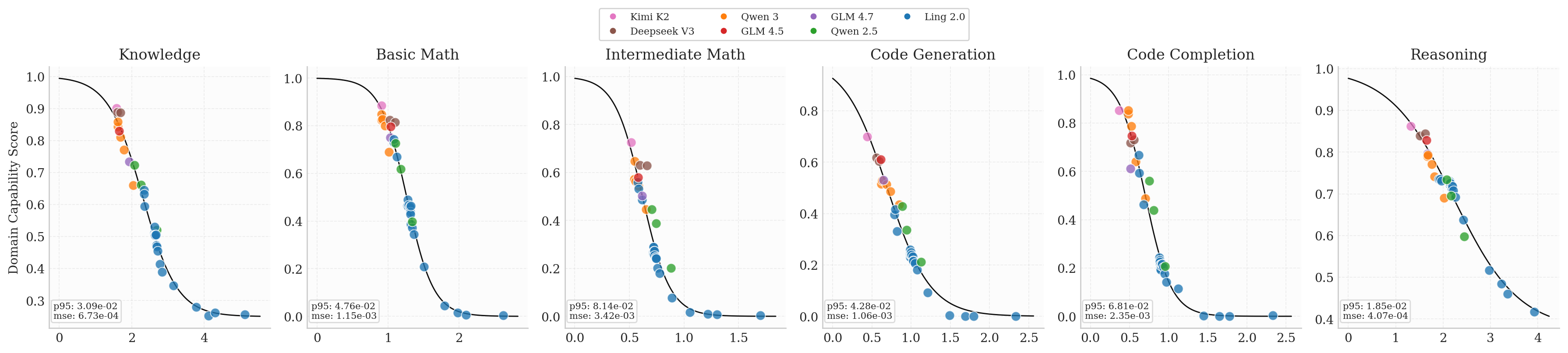} \hfill
  \caption {The domain-level loss and capability fitting results on open-source models. Each data point in a figure represents an open-source model with SuperValid loss and capability performance for that domain. Solid black line represents the sigmoid fitting curve. The results show that SuperValid is applicable on different domains, proving SuperValid can synthesize domain capability-aware validation data.}
  \label{fig:openModelSupervalid}

\end{figure*}

\subsection{SuperValid: Capability-Aware OOD Validation Data}
\label{sec:syntheticSteps}






\begin{algorithm}[t]
\caption{Capability-Aware OOD Data Synthesis for Domain $d_k$}
\label{alg:supervalid}
\begin{algorithmic}[1]
\Require Domain $d_k$, benchmark set $\mathcal{B}_k$, corpus $\mathcal{C}$
\Ensure Domain-specific validation set $\mathcal{V}_k$

\Statex \textit{// Initialization}
\State Validation set $\mathcal{V}_k \leftarrow \emptyset$, knowledge set $\mathcal{K}_k \leftarrow \emptyset$
\State Gather domain samples $\mathcal{S}_k \leftarrow \bigcup_{b \in \mathcal{B}_k} \mathcal{S}_b$

\Statex \textit{// Knowledge extraction from $\mathcal{S}_k$}
\For{$s \in \mathcal{S}_k$}
    \State $\mathcal{K}_k \leftarrow \mathcal{K}_k \cup f_M(s, d_k)$
\EndFor

\For{ Knowledge factor $q \in \mathcal{K}_k$}
\Statex \textit{// Retrieval}
    \State $\mathcal{R}_q \leftarrow f_R(q, \mathcal{C})$
\Statex \textit{// Filtering}
    \State $\tilde{\mathcal{R}}_q \leftarrow f_J(\mathcal{R}_q, q, d_k)$
\Statex \textit{// OOD generation}
    \State $\mathcal{V}_k \leftarrow \mathcal{V}_k \cup f_G(\tilde{\mathcal{R}}_q, q, d_k)$
\EndFor

\State \Return $\mathcal{V}_k$
\end{algorithmic}
\end{algorithm}

As evidenced by preliminary experiments (Figure~\ref{fig:compareIID}), IID validation loss is often insufficient for predicting downstream capability, as it primarily captures the likelihood under the training distribution~\citep{liu2023same} rather than domain-specific competence.
We therefore seek to construct OOD validation data with broader semantic coverage that better aligns with domain capabilities.


The key insight is that benchmark questions are only sparse instantiations of domain capability under fixed formats and scenarios~\citep{liang2022holistic}, and synthesizing from them directly risks preserving benchmark-specific artifacts.
SuperValid therefore targets the latent factors behind domain capabilities, including core knowledge concepts, reasoning patterns, and application scenarios.
Given benchmark questions from a domain $d_k$, it abstracts their underlying capability factors, retrieves and filters related content from a large corpus, and rewrites them into OOD validation samples that are less tied to surface-level formats and more reflective of domain-level capabilities.

Concretely, SuperValid consists of four stages as shown in Figure \ref{fig:supervalidFramework}:
\begin{itemize}
    \item \textbf{Knowledge Extraction.} An LLM $f_M$ extracts domain-relevant keywords, concepts, and reasoning patterns as knowledge factors that constitute the knowledge set $\mathcal{K}_k$ from benchmark questions $S_k$.
    \item \textbf{Knowledge Retrieval.} Each extracted knowledge factor $q$ in $\mathcal{K}_k$ is used to retrieve related candidate evidence $\mathcal{R}_q$ from a large corpus $\mathcal{C}$. This can be viewed as retrieving a common scenario for each factor.
    \item \textbf{Relevance Filtering.} An LLM-based judge $f_J$ removes weakly related retrieved content, retaining valid items in $\tilde{\mathcal{R}}_q$. An empty $\tilde{\mathcal{R}}_q$ is produced when all candidates are filtered out.
    \item \textbf{Scenario Expansion.} An LLM $f_G$ augments and rewrites $\tilde{\mathcal{R}}_q$ into diverse validation samples. Each resulting sample consists of detailed knowledge factors and questions under various scenarios in the form of questions, analyses, and solutions at different difficulty levels.
\end{itemize}
The resulting samples constitute the domain-specific validation set $\mathcal{V}_k$ for domain $d_k$. The cross-entropy loss of $\mathcal{V}_k$ is termed SuperValid loss.
 
The procedures above are summarized in Algorithm~\ref{alg:supervalid} using the aforementioned notations. To illustrate the synthesis process in a specific case, we present an example in Figure~\ref{fig:supervalidFramework}. Appendix~\ref{app:formalization} provides a detailed formulation, and all prompts for LLMs in the SuperValid algorithm are listed in Appendix~\ref{app:prompts}.

\begin{table*}[ht]
    \centering
    \caption{Statistics of SuperValid-synthesized validation data per domain. zhongkao2024\_math and gaokao2024\_math are in-house benchmarks collected from China's 2024 senior high school entrance examination (Zhongkao) and national college entrance examination (Gaokao), respectively.}
    \begin{tabularx}{\textwidth}{
    >{\hsize=0.8\hsize}L
    >{\hsize=0.4\hsize}L
    >{\hsize=0.5\hsize}L
    >{\hsize=2.3\hsize}X
    }    \toprule
     \textbf{Domain} & \textbf{Sample Counts} & \textbf{Average Sample Length} & \textbf{Benchmarks} \\ \midrule
Knowledge &	19813&4235.67 & CEval\citeyearpar{huang2023ceval}, CMMLU\citeyearpar{li2024cmmlu}, MMLU\citeyearpar{hendrycks2020mmlu} \\  
Basic Math&12572&1333.59 & MSGM\citeyearpar{shi2022mgsm}, CMath\citeyearpar{wei2023cmath}, zhongkao2024\_math \\ 
Intermediate Math&21711&1592.48 & CollegeMath\citeyearpar{tang2024collegemath}, gaokao2024\_math, MATH\citeyearpar{hendrycks2021math} \\ 
Code Generation&3871&4802.78  & MBPP\citeyearpar{austin2021mbpp}, MBPP\_PLUS, BIRD\_SQL\citeyearpar{li2023birdsql}\\ 
Code Completion&3237&2527.29 & HumanEval\_Plus\citeyearpar{liu2023humanevalplus}, HumanEval\citeyearpar{chen2021openaihumaneval}\\ 
Reasoning&24898&2319.16 & HellaSwag\citeyearpar{zellers2019hellaswag}, PIQA\citeyearpar{bisk2020piqa}\\ \bottomrule
    \end{tabularx} 
    \label{tab:domainBenchmarks}
\end{table*}

\subsection{Connecting SuperValid Loss to Downstream Capability}
\label{sec:scalingLaw}

With the synthesized validation set in hand, we now establish a quantitative link between validation loss and downstream capability.
For each domain $d_k$, let $\mathcal{V}_k=\{x_{k,1},\ldots,x_{k,n_k}\}$ denote the validation set produced by SuperValid. We define the domain validation loss of model $M$ as
\begin{equation}
    L_k(M) = \frac{1}{|\mathcal{V}_k|}
    \sum_{x\in\mathcal{V}_k} \ell(M,x),
\end{equation}
where $\ell(M,x)$ is the token-level cross-entropy loss on sample $x$.

Following prior work on loss-to-performance mapping~\citep{ge2025capability}, we connect $L_k(M)$ to domain capability via a bounded sigmoid:
\begin{equation}
    \label{eq:domainScalingLaw}
    f_k(L)
    =
    \gamma_k + \frac{1-\gamma_k}
    {1+\exp\left(\alpha_k(L-\beta_k)\right)} ,
\end{equation}
where $f_k(\cdot)$ is the predicted capability score for domain $d_k$, $\gamma_k$ is the chance-level performance lower bound, the upper bound is fixed at $1$, and $\alpha_k, \beta_k$ are domain-specific parameters controlling the curve's steepness and midpoint.
The sigmoid is monotonically decreasing in $L$, encoding the expectation that lower validation loss corresponds to stronger capability.

For each domain $d_k$, we fit the scaling function by minimizing the mean squared error between the predicted capability score and the benchmark-derived capability proxy:
\begin{equation}
    \min_{\alpha_k,\beta_k}
    \frac{1}{|\mathcal{M}|}
    \sum_{M\in\mathcal{M}}
    \left(
    \hat{c}_k(M) -
    f_k(L_k(M))
    \right)^2 ,
\end{equation}
where $\mathcal{M}$ is the set of models used for fitting, and $\hat{c}_k(M)$ is the observed domain capability proxy defined by aggregating benchmark scores within domain $d_k$.
Operating at the domain level rather than the individual benchmark level allows us to evaluate whether the validation set captures stable capability signals that generalize across benchmarks within the same domain.

\section{Experiment}

In this section, we verify that the SuperValid validation set possesses the properties described in previous sections. The main results (\S\ref{sec:openModelsFitting}) demonstrate that SuperValid loss faithfully reflects domain capabilities on various models, surpassing benchmark-level methods and IID loss. In the rest parts, we discuss our pipeline effectiveness (\S\ref{sec:pipelineAblation}) as well as robustness to training-data switching. (\S\ref{sec:midTrainComparison}).

\subsection{Settings}

\paragraph{Domains and Benchmarks.}
To conduct the experiments, we empirically select 16 benchmarks grouped into six domains for evaluation. The details of the benchmarks and domains are shown in Table \ref{tab:domainBenchmarks}.

\paragraph{SuperValid Settings.}
Using the SuperValid synthesis pipeline described in Algorithm \ref{alg:supervalid}, we construct a validation set containing 86,102 samples. The number of samples in each domain varies for different domains. Detailed sample features of domains can be found in Table \ref{tab:domainBenchmarks}. 

\begin{figure*}[t]
\centering
  \includegraphics[width=0.95\textwidth]{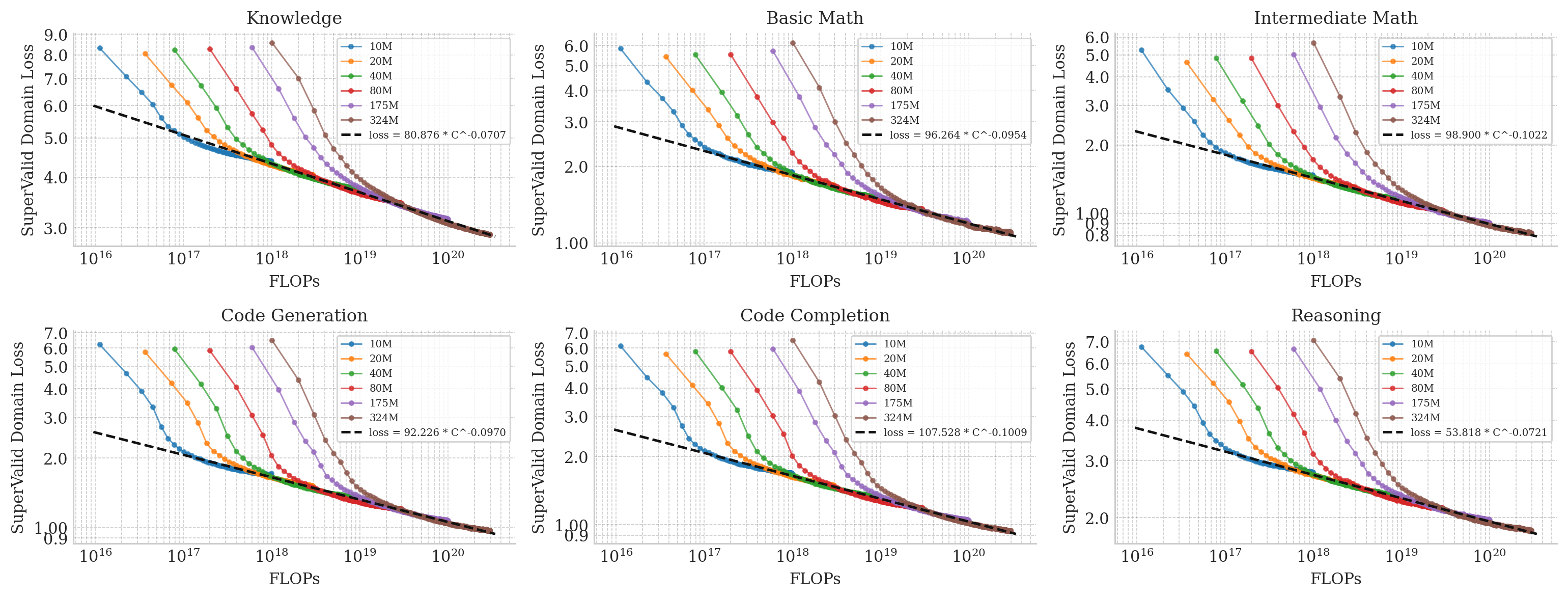} \hfill
  \caption {Scaling law curve on different model scales. Points and lines in different colors denote compute and SuperValid loss collected from models of different scales trained from scratch. The results demonstrate that SuperValid loss on different models largely obey the log-linear scaling law curve.}
  \label{fig:scalingLawCurve}

\end{figure*}

We utilize the FineWeb~\citep{penedo2024fineweb} dataset as the corpus. Retrieving text from a large corpus is a challenging task. To improve efficiency, we utilize OLMOTrace~\citep{liu2025olmotrace}, a high-performance text retrieval framework.
For the generative models used in the pipeline, Qwen3-Next-80B-A3B-Instruct~\citep{yang2025qwen3} is used for knowledge extraction and expansion. DeepSeek-V3.1-Terminus-Think~\citep{liu2024deepseek} is used to judge the relevance of retrieved text.

\paragraph{Models.}

All the models used in our experiments are open-weight models. For experiments in a training scenario, we refer to the model architecture of Ling-Mini-2.0-Base~\citep{li2025ling2} to create models with different scales. For training-free experiments, we test Qwen3~\citep{yang2025qwen3}, Kimi K2~\citep{team2025kimi}, GLM 4.5~\citep{zeng2025glm}, DeepseekV3~\citep{liu2024deepseek} etc. with their base models. Detailed models used in the experiments are shown in Appendix \ref{app:openModelsList}.

\paragraph{Evaluation Details.}
We use the OpenCompass~\citep{2023opencompass} tool to evaluate the downstream benchmark scores for all models in our experiments. The evaluation backend is SGLang~\citep{zheng2024sglang}. We set the inference context length to 8192 for alignment on different models. The temperature during inference is 0.

\paragraph{Fitting Evaluation Metrics.}
To fit the sigmoidal function, we utilize the L-BFGS-B~\citep{zhu1997L-BFGS-B} optimization method provided by Scipy~\citep{2020SciPy-NMeth} Python package. To evaluate the performance of the adjustment, we use the MSE and the width of the 95 percent confidence interval as metrics. This value is calculated as $P_{95} = 1.96\bar{R}$, where $R$ is the absolute standard variance of the residual errors. A smaller $P_{95}$ value indicates better predictive performance.

\subsection{Main results}

\paragraph{Domain-Level Capability Indicator for Open-Source Models.}
\label{sec:openModelsFitting}

To validate the SuperValid validation loss as a effective metric for evaluating domain-level capabilities, we use SuperValid data as a validation set for open-source models with six domains. Figure \ref{fig:openModelSupervalid} shows that the sigmoidal fitted curve can effectively predict the downstream capabilities for various open-source models on different domains with the MSE values within the range of 1e-3. These results demonstrate that SuperValid validation set is able to capture domain-specific capabilities across open-source models with different families, architectures, and scales, showing that SuperValid loss is a training-free and general downstream capabilities indicator.

\paragraph{Domain-Level Loss Improves Cross-Domain Prediction.}
\label{sec:crossDomainPrediction}

To prove the generalization performance of domain-level validation, we conduct a cross-domain prediction experiment on the aforementioned models. We synthesize a benchmark-level validation set with SuperValid by limiting the data source to a single benchmark of a domain.

\begin{figure}[t]

\centering
\includegraphics[width=\columnwidth]{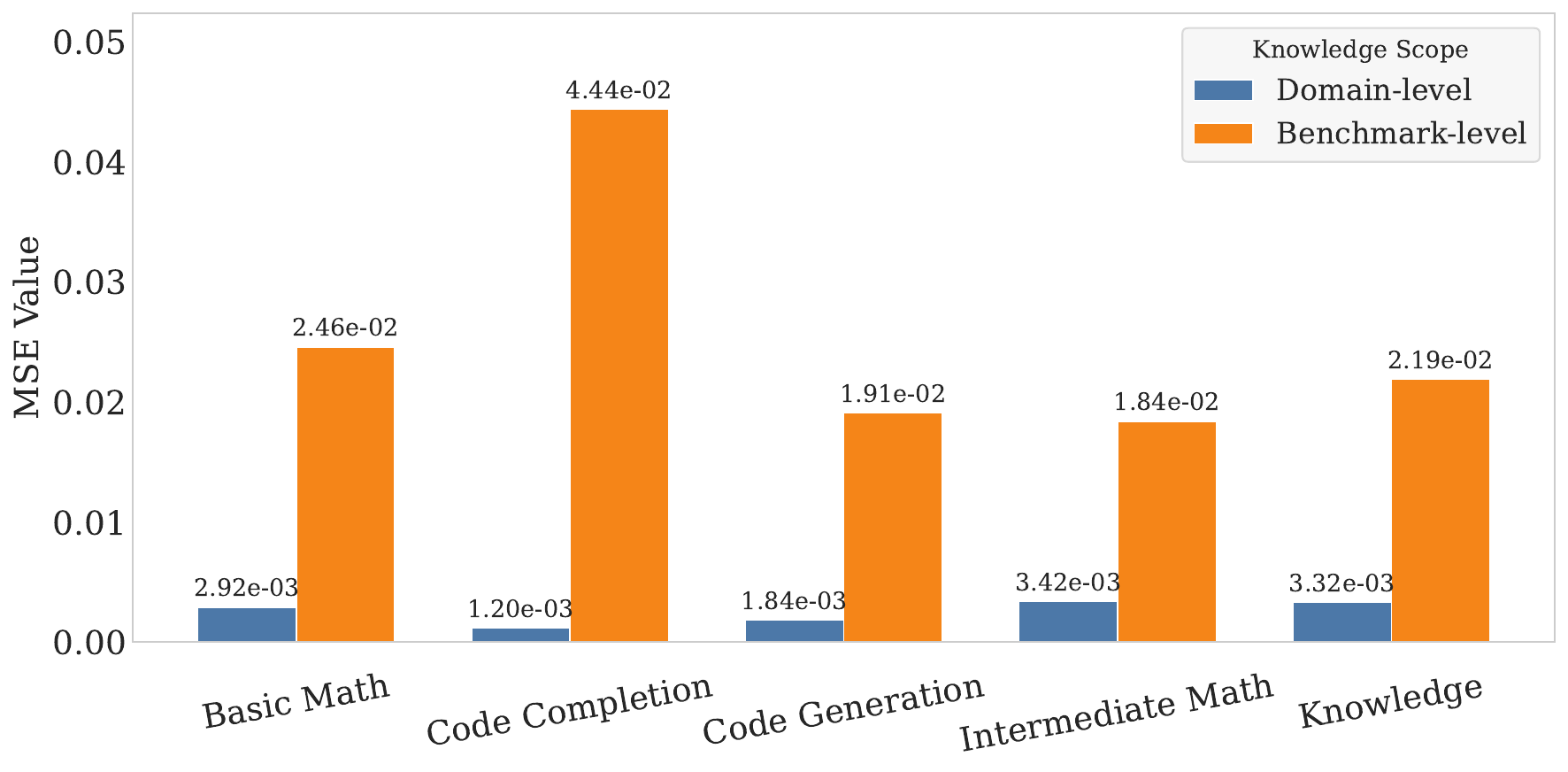}
\caption{Fitting results for domain capabilities with either domain-level loss or benchmark-level loss. The latter is derived from a validation set constructed by SuperValid with merely a single benchmark as data source. We instantiate the domain-level OOD set with \emph{Intermediate Math}, and benchmark-level OOD set with CollegeMath~\citep{tang2024collegemath} which is from the same domain. The MSE values in each group denote prediction error. The results show that domain-level loss captures downstream capability better than benchmark-level loss, exhibiting cross-domain generalization. }
\label{fig:domainlossfitting}

\end{figure}

\begin{figure}[t]

\centering
\includegraphics[width=\columnwidth]{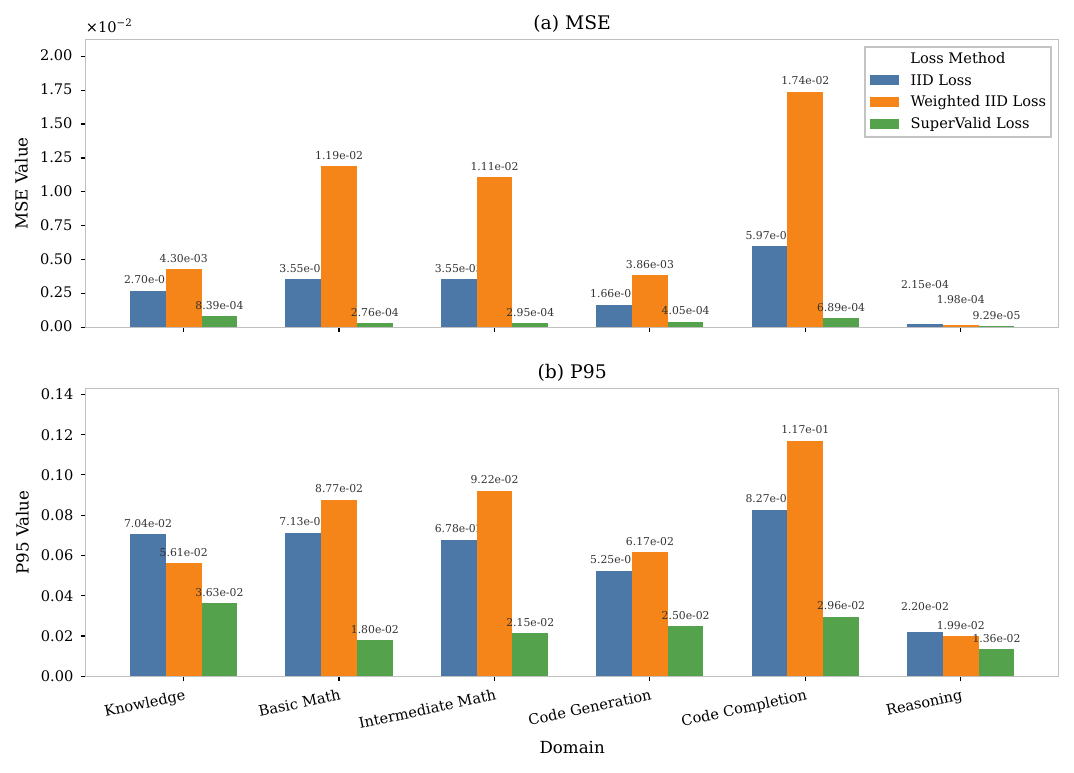}
\caption{Fitting results for domain capabilities with 3 different methods: IID loss, weighted IID loss and SuperValid loss. The results show that traditional IID-loss-based methods fail to capture the downstream capability while SuperValid performs better. }
\label{fig:ComparisonOnDomains}
\end{figure}

Figure~\ref{fig:domainlossfitting} shows that domain-level SuperValid loss provides a stronger predictor than benchmark-level loss in cross-domain settings. For both in-domain and cross-domain targets, domain-level loss achieves lower fitting errors, suggesting that it captures more generalizable capability signals rather than benchmark-specific patterns. This supports our motivation of modeling downstream scaling at the domain-capability level: domain-level loss offers a more stable and transferable proxy for downstream performance, which is crucial for reliable downstream scaling analysis across domains.

\begin{figure*}[t]
\centering
  \includegraphics[width=0.92\linewidth]{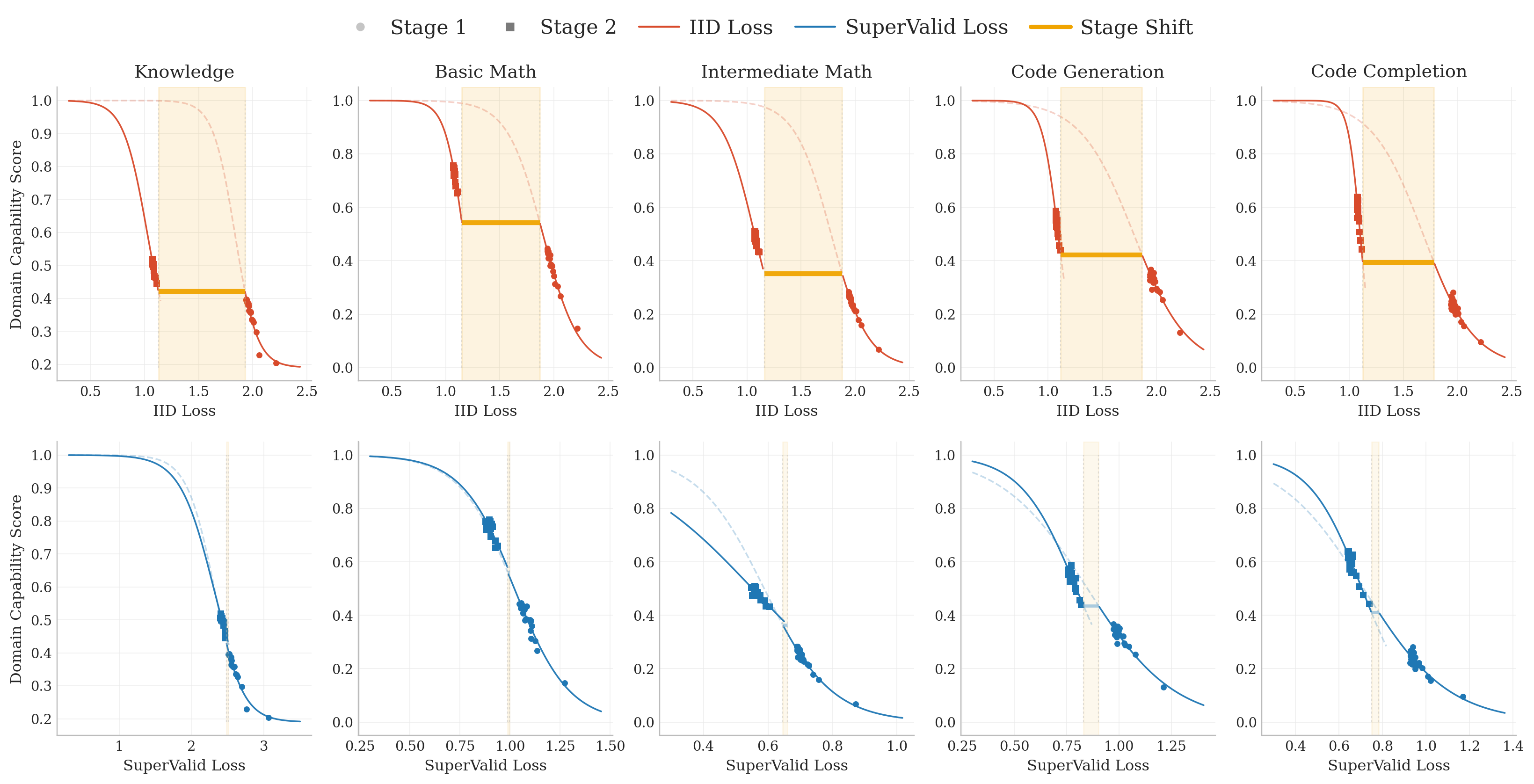}
  \caption{Effect of mid-training data switching on IID loss versus SuperValid loss. We fit a dedicated capability sigmoidal curve for each training stage and measure the gap at the transition point. When training data is switched in stage~2, IID loss (up) exhibits a large discontinuity between the two stages, whereas SuperValid loss (down) maintains a consistent loss-to-capability relationship.}
  \label{fig:dataScaling}
\end{figure*}

\paragraph{Leveraging Better Generalization with SuperValid.}
\label{sec:compareIID}

To validate the effectiveness of OOD validation compared to IID validation, we train 3 models with arbitrary architectures and data-mixtures, which is a common scenario for ablation studies. Figure \ref{fig:compareIID} shows the comparison of downstream task predicting on the CEval benchmark between IID loss and SuperValid loss. Figure \ref{fig:ComparisonOnDomains} shows the results with those methods at domain-level. We implement a prior token-level loss re-weight method from \citet{ge2025capability} and learn the re-weight mapping on two of the shown experiments. The results illustrate that SuperValid loss aligns better with downstream task performance, while IID loss fails to track the downstream performance with different training settings even with token-level loss re-weight. It is demonstrated that the SuperValid data have better generalization comparing with traditional IID data.

\paragraph{Log-Linear Scaling Law for SuperValid Loss.}
\label{sec:logLinearScalingLaw}

To investigate whether the property of log-linear scaling law can be observed on SuperValid loss, we test a series of MoE models with similar architectures but different scales with the active parameters from 10M to 324M and corresponding total paramerter from 193M to 6.17B. We train those models from scratch with the same training data. The learning rate varies from 1.73e-3 to 7.4e-4 as the model scale increases to align the training efficiency.
The results in Figure \ref{fig:scalingLawCurve} clearly show that the SuperValid loss follows a typical scaling law curve with a clear log-linear relationship, obeying the scaling law. This experiment is strong evidence that the SuperValid loss is suitable for the scaling law prediction method, providing a potential to achieve reliable downstream scaling on compute budget, model scales, and data consumption.

\subsection{From Surface Matching to Capability Alignment}
\label{sec:pipelineAblation}

To validate the effectiveness of our synthesis pipeline, we design the ablation experiments by skipping two key steps in our pipeline: knowledge extraction and knowledge expansion.

The experiment that skips knowledge extraction uses blank-reference filling, in which we fill the benchmark query text with correct answers directly as validation samples. We utilize the aforementioned benchmarks to generate 70660 samples in total, with an average sample length of 1582.86.

The knowledge expansion experiment skips the pipeline procedures after (iii), using the retrieved text from the corpus directly as the validation samples. Due to the huge number of retrieved text samples, we randomly choose 80000 samples to align the quantity with the other two experiments.

The MSE results in Table \ref{tab:ablationResult} reveal that simply matching the surface form of the benchmark text (blank filling) is insufficient to capture downstream capability. Introducing retrieval improves alignment by incorporating semantically relevant content, but still falls short due to limited coverage and lack of contextual diversity. 
SuperValid further improves performance by expanding and rewriting knowledge into diverse contexts, resulting in a validation distribution that better captures the underlying capabilities required by downstream tasks.
These results suggest that effective validation signals require not only semantic relevance, but also sufficient diversity and contextualization, which are achieved through knowledge synthesis. 

\begin{table}[ht]
    \centering
    \caption{Synthesis method ablation on knowledge domain of open-source models. As the synthesis pipeline approaches the full SuperValid pipeline, the MSE and $P_{95}$ values decrease, indicating better alignment. This proves the effectiveness of the SuperValid pipeline.}
    \begin{tabularx}{\columnwidth}{>{\hsize=1.2\hsize}L>{\hsize=0.9\hsize}L>{\hsize=0.9\hsize}L}    \toprule
     \textbf{Method} & \textbf{MSE} & $\mathbf{P_{95}}$ \\ \midrule
     SuperValid & \textbf{6.73e-4} & \textbf{3.09e-2} \\ 
     Retrieved Text & 2.31e-3 & 5.17e-2 \\ 
     Blank Filling & 7.29e-3 & 1.17e-1 \\ \bottomrule
    \end{tabularx} 
    \label{tab:ablationResult}
\end{table}

\subsection{Train Data Switching Robustness}
\label{sec:midTrainComparison}

Mid-training has been increasingly adopted in pre-training practices, where different data mixtures are used at different stages to meet evolving training requirements~\citep{mo2025mid}. Although effective, this method of shifting the training distribution inevitably causes a significant loss shift in IID loss. However, it is not the case for SuperValid loss.

We set up an experiment to train a model from scratch with a total number of 2T(2e12) tokens, divided into two stages of 1T tokens each. During training, we collect both IID validation loss and SuperValid loss every 25B(2.5e10) tokens.
For stage 1, a data mixture with a small proportion of synthetic data is applied. After the training process reaches stage 2, we change the data mixture by increasing the proportion of synthetic data.
This setup allows us to evaluate how validation signals respond to distribution shifts during training.

The experiment results in Figure \ref{fig:dataScaling} show that switching data mixtures during training causes a tremendous shift in IID loss, which disrupts traditional scaling-law tracking, leading to disagreement between scaling law and mid-training. Meanwhile, SuperValid loss manifests good robustness when switching data mixtures, allowing downstream scaling laws to remain compatible with mid-training.
This confirms that IID loss is sensitive to changes in training data distribution and fails to provide a consistent proxy for downstream performance during mid-training. This robustness to data distribution shifts allows SuperValid to recover a consistent scaling behavior, making it a reliable signal for tracking capability.

\section{Related Work}

\paragraph{Scaling Laws and Performance Prediction. }
Scaling laws~\citep{bahri2024explaining, hoffmann2022training,kaplan2020scaling,muennighoff2023scaling} have been widely used to guide LLM training. Recent work extends scaling analysis to downstream performance prediction, such as modeling benchmark scores from compute~\citep{chen2024scaling}, model scale and pretraining loss~\citep{hu2023predicting}. Another line of work characterizes model capabilities through low-dimensional structures derived from benchmark matrices, using principal component analysis~\citep{ruan2024observational}, factor analysis~\citep{burnell2023revealing}, or latent-variable models~\citep{jin2025discovering,zhang2024collaborative}. However, these approaches all operate on benchmark scores and do not address how to obtain better validation signals from the loss side.

\paragraph{Validation Signals for Capability Estimation.} 
Traditional validation sets are constructed from training data subsets, and the corresponding loss is termed IID validation loss~\citep{tamang2025handling}. Recent works~\citep{liu2023same,lourie2025unreliabledownstream,teney2023id,isik2024scaling, mckenzie2022inverse} show that differences in downstream performance across pretraining corpora are often only weakly reflected in IID validation loss. Several methods attempt to improve validation signals without changing the underlying data, such as learning token-level importance weights~\citep{ge2025capability}, data-mixture-aware loss proxies~\citep{li2026capacity}, capability-specific skill representations~\citep{maia2026sloth,ye2023flask}, benchmark metrics decomposition~\citep{ruan2024observational}, analysis of emergent abilities~\citep{du2024understanding}, and proxy model evaluation~\citep{anugraha2025proxylm,koh2025predicting}. In contrast, SuperValid improves the validation signal by constructing new OOD validation data rather than reweighting existing data or metrics.

\paragraph{OOD Data Generation. }
OOD data have been previously investigated in fields such as OOD detection in CV \citep{yu2024survey}. Some of those works focus on OOD data generation \citep{yang2024generalized,tamang2025handling}, which provide insights for the NLP field on evaluation and generalization~\citep{li2023survey,yang2023out}. For instance, \citet{cao2024envisioning} provide an idea of synthesizing OOD data using LLMs. \citet{abbas2025out} propose a framework to synthesize structured OOD text data with different levels of distribution shifting strength using LLMs. \citet{lin2026oodbench} and \citet{koh2021wilds} constructs OOD benchmarks for better evaluation on model capabilities. \citet{zhou2020learning} proves that testing on domain-level OOD data is benefitial for model generalization. \citet{feder2023data} involves counterfactual data augmentation to strengthen OOD generalization. Those works inspire us to further utilize OOD data as validation set to achieve downstream capability prediction with better generalization.

\section{Conclusion}

In this work, we revisit the role of validation in large language model training and show that conventional IID validation loss is an unreliable proxy for downstream capability, particularly under distribution shifts and across model scales. To address this limitation, we propose SuperValid, a synthetic validation framework that constructs a capability-aligned validation distribution by distilling and expanding knowledge from domain-clustered downstream benchmarks without involving benchmark-specific artifacts. Extensive experiments demonstrate that SuperValid loss exhibits a strong and consistent correlation with domain capabilities across different model architectures and scales, and also remains stable under data-distribution changes, indicating a more faithful signal for model selection, early stopping, and scaling decisions.




\section*{Limitations}

First, SuperValid remains benchmark-informed rather than fully benchmark-independent. Domain capability is approximated by an unweighted average of manually grouped benchmark scores, while the validation data are seeded from the same benchmark collection. Although our synthesis pipeline reduces dependence on benchmark-specific formats and surface patterns, it may still inherit biases in capability definition and knowledge coverage from the selected benchmarks.

Second, cross-family SuperValid losses are computed using each model's native tokenizer. Differences in tokenization granularity may therefore affect the comparability of per-token cross-entropy across model families. Future work should consider byte- or character-normalized likelihoods to better isolate capability-related differences.

The optimal validation-set size for capability evaluation has not been studied in our work. In our experiments, we generate approximately 80,000 samples in total. The size of validation samples implies a trade-off between validation time during training and effectiveness of capability prediction. Therefore, the sweet spot of synthetic validation set size is still an interesting topic to be explored. Another unexplored area of our work is that we select the benchmarks for domains based on empirical knowledge and community consensus, and additional domains, such as multilinguality or security, are not included. Nevertheless, the selected domains provide sufficient evidence to support our claims; exploring additional domains and benchmarks remains future work.



\bibliography{custom}

\appendix

\section{Formalization of SuperValid Synthesis}
\label{app:formalization}

We provide a formal description of the SuperValid synthesis pipeline introduced in Algorithm~\ref{alg:supervalid}. 
For a capability domain $d_k$, let $\mathcal{B}_k=\{b_{k,1},\ldots,b_{k,n_k}\}$ denote the set of benchmarks associated with this domain. 
Each benchmark $b\in\mathcal{B}_k$ contains a set of original benchmark samples $\mathcal{S}_b$. 
We first collect all benchmark samples within the domain:
\begin{equation}
    \mathcal{S}_k = \bigcup_{b\in\mathcal{B}_k} \mathcal{S}_b .
\end{equation}

SuperValid synthesis process is defined by four functions: knowledge extraction $f_M$, retrieval $f_R$, relevance filtering $f_J$, and sample generation $f_G$.

First, for each benchmark sample $s\in\mathcal{S}_k$, we use an LLM-based extractor $f_M$ to identify domain-relevant knowledge factors:
\begin{equation}
    \mathcal{K}_k
    =
    \bigcup_{s\in\mathcal{S}_k}
    f_M(s,d_k),
\end{equation}
where $\mathcal{K}_k$ denotes the set of extracted knowledge factors for domain $d_k$.

For each knowledge factor $q\in\mathcal{K}_k$, we retrieve a set of relevant text from a large corpus $\mathcal{C}$:
\begin{equation}
    \mathcal{R}_q = f_R(q,\mathcal{C}).
\end{equation}
Since retrieval may introduce noisy or weakly related text, we further apply an LLM-based judge $f_J$ to select texts that are relevant to both the knowledge factor $q$ and the domain $d_k$:
\begin{equation}
    \tilde{\mathcal{R}}_q = f_J(\mathcal{R}_q,q,d_k),
\end{equation}
where $\tilde{\mathcal{R}}_q \subseteq \mathcal{R}_q$ is the filtered evidence set.

Finally, the generator $f_G$ rewrites the filtered evidence into validation samples. 
Each generated sample may contain contextual knowledge, questions, analyses, and solutions designed to probe the corresponding domain capability:
\begin{equation}
    \mathcal{V}_{k,q}
    =
    f_G(\tilde{\mathcal{R}}_q,q,d_k).
\end{equation}
The final domain-specific validation set is obtained by aggregating generated samples over all extracted knowledge factors:
\begin{equation}
    \mathcal{V}_k
    =
    \bigcup_{q\in\mathcal{K}_k}
    \mathcal{V}_{k,q}
\end{equation}

By retrieving, filtering, and expanding knowledge around these factors, SuperValid constructs $\mathcal{V}_k$ as a capability-aware OOD validation set with broader semantic coverage than benchmark-specific or IID validation data.


\section{Open-Source Model List}
\label{app:openModelsList}

Table~\ref{tab:openModelsList} provides the open-source models used in our experiments. We get the model weights and standard configurations from their Hugging Face pages.

\begin{table}[h]
    \centering
    \caption{Open-source models for the experiments.}
    \label{tab:openModelsList}
    \begin{tabularx}{\columnwidth}{>{\hsize=0.8\hsize}L>{\hsize=1.2\hsize}L}    \toprule
     \textbf{Family} & \textbf{Model}  \\ \midrule
     Qwen3 & Qwen3-1.7B-Base Qwen3-4B-Base Qwen3-8B-Base Qwen3-14B-Base Qwen3-30B-A3B-Base \\ 
     \midrule
     Qwen2.5 & Qwen2.5-0.5B-Base Qwen2.5-1.5B-Base Qwen2.5-3B-Base \\ \midrule

     GLM4.5 & GLM-4.5-Air-Base \\ \midrule
     GLM4.7 & GLM-4.7-Flash-Base \\ \midrule
     DeepSeekV3 & DeepSeek-V3-Base DeepSeek-V3.1-Base \\ \midrule
     Kimi-K2 & Kimi-K2-Base \\ \midrule
     Ling-2.0 & Ling-Mini-Base-2.0 \\ \bottomrule

    \end{tabularx} 
    \label{tab:openSourceModelList}
\end{table}

\section{Synthesis Prompt}
\label{app:prompts}

The prompts applied in SuperValid in our experiments are shown in this section. We provide three prompts corresponding to three LLM-required procedures of SuperValid, including knowledge extraction, relevance filtering, and scenario expansion. The template form \emph{\$\{var\}} should be filled with the actual text data such as benchmark text, extracted concepts, etc.

\begin{figure*}[t] 
\centering
\begin{tcolorbox}[
  colback=blue!3!white,
  colframe=blue!60!black,
  boxrule=1.5pt,
  arc=4pt,
  left=8pt,
  right=8pt,
  top=6pt,
  bottom=6pt,
  width=0.96\textwidth,
  title={\textbf{Prompts for Knowledge Extraction}},
  fonttitle=\bfseries,
  coltitle=white,
  label={app:prompt_knowledge_ext},
]
\small

\textbf{You are a professional knowledge search engine optimization expert}. Your task is to optimize a teacher's lesson plan question into the most searchable single knowledge concepts, entities, or phrases in \textbf{keyword form}, such that correctly answering the question requires mastery of these knowledge keywords.

\medskip
\textbf{Task Requirements}:

\begin{enumerate}
\item \textbf{Core Intent Knowledge Extraction}: Capture the core knowledge of the question. This can be a summary of the question's knowledge or the extraction of key knowledge keywords from it. You must ignore redundant, descriptive, colloquial modifiers and purely numerical descriptions.
\item \textbf{Concept Decomposition}: Break down composite concepts into more fundamental keywords that are more likely to appear as titles in knowledge documents, making them suitable as **search knowledge keywords**.
\item \textbf{Output Format}:
\end{enumerate}

\begin{itemize}
\item Knowledge keywords should use precise and concise terminology, e.g., "Newton's Second Law," "Ideal Gas Equation of State," "Law of Diminishing Marginal Utility," "DNA Semiconservative Replication."

\item Each knowledge keyword should represent a complete and targeted knowledge concept that covers a full concept, e.g., "Reflection of Light" instead of "Reflection" or "Light," "1990 Billboard Year-End Chart" instead of "1990" or "Year-End Chart," "T cell" instead of "T."

\item Prohibited invalid knowledge keywords:

\begin{itemize}
    \item Vague or incomplete meanings, such as "calculate," "understand the situation," "boundary," "D," etc.
    \item Overly general or non-specific meanings, such as "water," "fire," "human," "she," "doubling time," and especially pure data types such as "100," "ten meters," "1990," "one million dollars," "128-bit," "400 meters or below," "sum of points equals 4," "[1]," "interval 1," etc.
\end{itemize}

\item Use standardized terminology.

\item Do not use Markdown or complex special symbols. Only use plain text and simple symbols.

\item The final result should be a string, separated by English commas (",").

\item Output at most 6 knowledge keywords.

\item Do not output any additional explanations. The output structure must strictly follow the format below:

  Extraction of key knowledge words:\\
  1. xxx\\
  2. xxx\\
  ......\\

\end{itemize}

\medskip
\textbf{Example}:

\medskip
\textbf{Lesson Plan Question}:
\medskip

"Chemistry is closely related to daily life. Which of the following statements is incorrect?\\
A. Using fluoride toothpaste can prevent dental caries\\
B. The main component of baking soda is Na2CO3\\
C. Vinegar can be used to remove calcium carbonate scale\\
Answer: B"\\

\medskip
\textbf{Output}:
\medskip

Extraction of key knowledge words:\\
1. Chemistry in daily life\\
2. fluoride toothpaste\\
3. prevention of dental caries\\
4. baking soda\\
5. vinegar\\
6. dissolving calcium carbonate\\

\medskip

\textbf{Lesson Plan Question}:
\medskip

\$\{raw\_exam\}
\medskip
\end{tcolorbox}
\end{figure*}

\begin{figure*}[t] 
\centering
\begin{tcolorbox}[
  colback=yellow!3!white,
  colframe=yellow!60!black,
  boxrule=1.5pt,
  arc=4pt,
  left=8pt,
  right=8pt,
  top=6pt,
  bottom=6pt,
  width=0.96\textwidth,
  title={\textbf{Prompts for Relevance Filtering}},
  fonttitle=\bfseries,
  coltitle=white,
  label={app:promptRelvanceFilter},
]
\small

Please carefully read, understand, and reason step by step based on the knowledge concept and knowledge learning text provided below, and \textbf{determine whether the knowledge learning text is strictly related to the knowledge concept}.

\medskip

Strict relevance criteria: Either the knowledge learning text directly contains descriptions of the knowledge concept and its related concepts/entities, or the knowledge learning text enables learning and understanding of the knowledge concept. Note that some knowledge entities and concepts may have aliasing, inclusion, or inferable relationships; such differences in expression can be ignored.
\medskip

Output format requirements: Do not output any redundant explanations. Directly follow the format below and include the judgment result inside [] brackets.
 
The judgment result can only be Yes or No:
\medskip

<Knowledge\_Concept\_Start>

\$\{knowledge\_concept\}

<Knowledge\_Concept\_End>
\medskip

<Candidate\_Relevant\_Text\_Start>

\$\{candidate\_retrieved\_content\}

<Candidate\_Relevant\_Text\_End>
\medskip

\textbf{Judgment Result:}

\textbf{Output}:
\end{tcolorbox}
\end{figure*}

\begin{figure*}[t] 
\centering
\begin{tcolorbox}[
  colback=blue!3!white,
  colframe=blue!30!white,
  boxrule=1.5pt,
  arc=4pt,
  left=8pt,
  right=8pt,
  top=6pt,
  bottom=6pt,
  width=0.96\textwidth,
  title={\textbf{Prompts for Scenario Expansion}},
  fonttitle=\bfseries,
  coltitle=white,
  label={app:promptRelvanceFilter},
]
\small

\textbf{You are an education expert}. Based on the knowledge material provided below, please carefully read, understand, and reason step by step, then complete the following tasks in order as required:
\medskip

\textbf{Step 1. Key Knowledge Concepts Analysis}:
\medskip

Understand the knowledge material, identify core knowledge concepts, conceptual relationships, logical reasoning scenarios, knowledge application scenarios, etc., and comprehend the key knowledge concepts.

\begin{itemize} 
\item Extract the core text and content related to key knowledge concepts.
\item Extract key action nodes and causal relationships in event chains.
\item Identify professional terminology and conceptual relationships.
\item Identify potential branching points in situational logic.
\item Analyze implicit application scenarios involving physical/social common sense.
\item The knowledge concept should be as specific and detailed as possible.
\end{itemize}

Output the key knowledge concepts in the following format:\\
Key Knowledge Concepts:\\
1. xxx\\
2. xxx\\
......\\
\medskip

\textbf{Step 2. Knowledge Expansion}:
\medskip

Based on the extracted key knowledge concepts, combined with the original material and your in-depth knowledge, expand the breadth and depth of the knowledge concepts and output related expanded knowledge in the following format:

Related Knowledge Expansion\\
1. xxx\\
2. xxx\\
......\\
\end{tcolorbox}
\end{figure*}

\begin{figure*}[t] 
\centering
\begin{tcolorbox}[
  colback=blue!3!white,
  colframe=blue!30!white,
  boxrule=1.5pt,
  arc=4pt,
  left=8pt,
  right=8pt,
  top=6pt,
  bottom=6pt,
  width=0.96\textwidth,
  title={\textbf{Prompts for Scenario Expansion}},
  fonttitle=\bfseries,
  coltitle=white,
  label={app:promptRelvanceFilter},
]
\small

\textbf{Step 3. Practice Generation}:
\medskip

Based on the core paragraphs of the knowledge material and combined with the key concepts, design and generate training questions in a targeted manner. The question type is limited to multiple-choice questions only. Note that the question design must follow the specifications below:
\medskip

Question Requirements:

\begin{itemize}
\item Principle: The questions must be based on the original material, but must not directly copy it!
\item Quantity: Generate as many questions as possible. At minimum, provide 5 questions strictly aligned with the key knowledge concepts, with a maximum of 10; for the expanded related knowledge concepts, provide at least 5 questions, with a maximum of 10.
\item Approach: The questions must be targeted toward knowledge concepts, not vague open-ended questions. The questions should focus on the knowledge concepts extracted from the original material, and must concern the knowledge concepts themselves while remaining independent of the original text content. Again, do not copy directly!
\item Diversity: The questions should be diverse and multi-dimensional. For the same concept, include different assessment perspectives such as knowledge recall, logical reasoning, computational proof, conceptual relationship discrimination, etc.
\item Difficulty: Include multiple levels of difficulty, such as foundational, professional, and competition-level questions.
\end{itemize}




Answer Specifications:

\begin{itemize} 
\item Ensure the standard answer option is correct.
\item You may appropriately include solutions and analyses for the questions, such as question interpretation, knowledge discrimination, method explanations, step-by-step reasoning, logical inference, viewpoint discussion, mathematical calculation processes, etc.
\end{itemize}
\medskip

Output Specifications:

\begin{itemize} 
\item Use plain text only. Do not use Markdown formatting.
\item Separate each question, options, answer, and analysis pair using line breaks. Use standard uppercase English letters such as A/B/C/D/E/F/G for options, and mark answers with “Answer:”.
\end{itemize}
\medskip

\textbf{Example Format}:
\medskip

<Question\_1\_Start>

xxx (question content)

A. xxx

B. xxx

C. xxx

D. xxx

(or more options)

Answer: xxx (answer content, including the answer option letter)

Analysis: xxx (analysis content)

<Question\_1\_End>

\medskip

Below is the provided knowledge material. Please begin the task.
\medskip

<Knowledge\_Material\_Start>

\$\{content\}

<Knowledge\_Material\_End>
\medskip

\textbf{Output}:
\end{tcolorbox}
\end{figure*}

\section{Synthesis Example}
\label{app:exampleCases}

To better illustrate the SuperValid pipeline, we show a canonical example from the knowledge domain. The results for every synthesis step are merged in the same text box with some omissions for better presentation.

\begin{figure*}[t] 
\centering
\begin{tcolorbox}[
  colback=red!3!white,
  colframe=red!60!black,
  boxrule=1.5pt,
  arc=4pt,
  left=8pt,
  right=8pt,
  top=6pt,
  bottom=6pt,
  width=0.96\textwidth,
  title={\textbf{An Example of SuperValid from Knowledge Domain}},
  fonttitle=\bfseries,
  coltitle=white,
  label={app:promptRelvanceFilter},
]
\small

\textbf{Source Data}\\

An obese individual with a breathing-related sleep disorder most likely suffers from which of the following?\\
A. Narcolepsy\\
B. Hypersomnia\\
C. Insomnia\\
D. Sleep apnea\\
Answer: D\\
\medskip

\textbf{Key Knowledge Concepts}\\

1. \textbf{Breathing-related sleep disorder}\\
2. Obstructive sleep apnea (OSA) is a sleep disorder characterized by recurrent pauses in breathing during sleep, primarily caused by mechanical obstruction of the airway due to relaxation of the muscles at the back of the throat.\\
3. Narcolepsy is a condition that makes people very sleepy during the day and can cause them to fall asleep suddenly.\\
......\\
\medskip

\textbf{Retrieved Content}\\

Depression, Anxiety, and Obstructive Sleep Apnea.\\
Sleep problems are a common reason that people see their doctor ...... Up to 20\% of people coming to their doctor with a \textbf{breathing-related sleep disorder} or a major depression disorder ......\\
\medskip

\textbf{Related Knowledge Expansion}\\

1. OSA Definition and Mechanism: Sleep apnea lasting $\ge$ 10 seconds, caused by relaxation of the pharyngeal muscles leading to mechanical airway obstruction.\\
2. The three main clinical signs of OSA are: nocturnal wheezing/breath-holding, snoring, and daytime sleepiness.\\
3. OSA by gender and age distribution: More prevalent in males, but present in all populations.\\
......\\
\medskip

\textbf{Questions}\\
\medskip

<Question\_1\_Start>\\
1. Which of the following combinations of signs most typically suggests that a patient may have obstructive sleep apnea (OSA)?\\

A. Nighttime chest pain, palpitations, daytime anxiety\\
B. Snoring, nighttime awakenings, daytime sleepiness\\
C. Persistent headache, blurred vision, agitation\\
D. Insomnia, early awakening, decreased appetite\\
Answer: B\\

Analysis: Key knowledge 2 and the extended information clearly list the three main signs of OSA: snoring, nighttime breathing difficulties/awakenings, and daytime sleepiness. B is a perfect match. A is a cardiovascular symptom, C is related to intracranial pressure, and D is a typical depressive symptom, but not a characteristic manifestation of OSA; it is necessary to distinguish between primary and secondary OSA.\\
<Question\_1\_End>\\
\medskip

<Question\_2\_Start>\\
2. Which of the following statements about the epidemiology of OSA is accurate?\\

A. OSA only occurs in obese men over 40 years of age.\\
B. OSA is extremely rare in women, and clinical screening of female patients is unnecessary.\\
C. OSA can occur in any sex and age, but the incidence is higher in men.\\
D. Childhood OSA is mainly caused by psychological stress and is unrelated to obesity.\\
Answer: C\\

Analysis: The extended information clearly states that "OSA is more common in men, but exists in all sexes and age groups," and C completely replicates this statement. A and B negate other groups, so they are incorrect; D is incorrect, as the main cause of childhood OSA is often adenoid hypertrophy, not psychological stress.\\
<Question\_2\_End>\\

\end{tcolorbox}
\end{figure*}

\end{document}